\def\year{2018}\relax
\documentclass[letterpaper]{article} %DO NOT CHANGE THIS
\usepackage{aaai18}  %Required
\usepackage{times}  %Required
\usepackage{helvet}  %Required
\usepackage{courier}  %Required
\usepackage{url}  %Required
\usepackage{graphicx}  %Required
\usepackage{amsmath}
\usepackage{amsthm}
\usepackage{amsfonts}
\usepackage[ruled]{algorithm2e}
\begin{document}

\newcommand*\diff{\ d}
\newcommand \pp{P}
\renewcommand{\thefootnote}{\fnsymbol{footnote}}
\title{Hierarchical Approaches for Reinforcement Learning \\in Parameterized Action Space}
\author{Ermo Wei \and Drew Wicke \and Sean Luke\\
Department of Computer Science, George Mason University, Fairfax, VA USA\\
ewei@cs.gmu.edu,\ \ dwicke@gmu.edu,\ \ sean@cs.gmu.edu
}
\maketitle
\begin{abstract}
We explore Deep Reinforcement Learning in a parameterized action space. Specifically, we investigate how to achieve sample-efficient end-to-end training in these tasks. We propose a new compact architecture for the tasks where the parameter policy is conditioned on the output of the discrete action policy. We also propose two new methods based on the state-of-the-art algorithms Trust Region Policy Optimization (TRPO) and Stochastic Value Gradient (SVG) to train such an architecture. We demonstrate that these methods outperform the state of the art method, Parameterized Action DDPG, on test domains.
\end{abstract}

\section{Introduction}
% RL have great success
% deep RL have not be extensive studied, but important
% We proposed two methods for solving the issue allow us to do end-to-end training
Deep Reinforcement Learning (DRL) has achieved  success in recent years, including beating human masters in Go~\cite{Silver2016Mastering}, attaining human level performance in Atari games~\cite{Mnih2015Human}, and controlling robots in high-dimensional action spaces~\cite{Lillicrap2015Continuous}. With these successes, researchers have begun to explore new frontiers in DRL, including how to apply DRL in complex action spaces. Consider for example the real time strategy game StarCraft, where at any time during play we may choose among different types of actions to be able to finish our goals~\cite{Vinyals2017Starcraft}. For example, we may need to choose a building to construct and then select where to build it; or choose a squad of armies and direct them to explore the map. Critically, instead of having a single action set, we may have several sets of actions, either continuous or discrete, and to get a meaningful action to execute, we must choose wisely among these sets. 

In this paper, we explore how to apply DRL to tasks with more than one set of actions. Specifically, we consider tasks with parameterized action spaces~\cite{Masson2016Reinforcement}, where at each step the agent must choose both a discrete action and a set of continuous parameters for that action. Tasks with this kind of action space have been proposed in the Reinforcement Learning (RL) community for a long time~\cite{Stone2005Keepaway} but have not been explored much. 

One approach to handle a RL task with a parameterized action space is to do {\it alternating optimization}~\cite{Masson2016Reinforcement}. Here, we break the task into two separate subtasks by fixing either the parameters or discrete actions and then applying RL algorithms alternating between the induced subtasks. Although this method can work, it has a huge sample complexity because every time we switch the subtask, the previous experience is no longer valid as the environment is different. 

Thus, a sample efficient alternative is to train the policies for discrete actions and parameters at the same time.  There have already been steps in this direction. ~\citeauthor{Hausknecht2015Deep} simultaneously train two policies which can produce the values for discrete action and parameters respectively and then select the action to execute based on their values. There are two main drawbacks of this method. The first is that the parameter policy does not know what discrete action is selected at execution time. Thus, the parameter policy needs to output all the parameters for all the discrete actions at every step. As a result, the output size of the parameter policy can explode if we have high dimensional parameters with large discrete action sets. The second problem is that neither the policies nor the training method are aware of the action-selection procedure after the action and parameter values are produced. Therefore, the method may be missing a crucial piece of information for it to succeed.

In this paper, we propose a new architecture for parameterized action space tasks. In our method, we condition our parameter policy on the output of the discrete action policy, thus greatly reducing the output size of the parameter policy. Then we extend the state-of-the-art DRL algorithms to efficiently train the new architecture for parameterized action space tasks.  In experiments we show that our methods can achieve better performance than the state of the art method.

%Although the independent assumption of policies for discrete and continuous actions allow us to do jointly training of the parameters, it doesn't really achieve what we want. For example, in soccer, a player may first make the choice of passing the ball, then look for the teammate to pass. In here, we can think of passing as the action, and whom to pass is the parameter for that action. It does not make sense of choosing the dribbling as the action then choose the teammate to dribble to. Thus, in these situations, we can see a hierarchy in the decision making, that is, we first decide the action and then decide the parameters for the action. Or more precisely, \(\pi(a,x|s) = \pi^{c}(x|a,s) \pi^{d}(a|s)\) as a variant of variational auto-encoder, where we first generate some hidden representation from the input and then generate a sample from the hidden representation. We denote \(\theta_a\) as the parameters for discrete action network, and \(\theta_x\) as the parameters for the continuous action network. In addtion, we define \(\Theta\) as the combination of both.
\section{Background}
Before we delve into the architecture and algorithms, we first present a mathematical formulation of Markov Decision Processes (MDPs) along with some relevant policy gradient algorithms. Then we present Parameterized Action MDPs (PAMDPs). Lastly, we discuss some of the previous work in PAMDPs that is related to our paper.
\subsection{MDPs and Policy Gradient Methods}
\paragraph{Markov Decision Process} A Markov Decision Process (or MDP) can be used to model the interaction an agent has with its environment.  A MDP is a tuple \(\{S, A, T, R, \gamma, H\} \) where  \(S\) is the set of states; \(A\) is the set of actions available to the agent; \(T\) is the transition function \(T(s, a, s') = {\pp}(s'|s,a)\) defining the probability of transitioning to state \(s'\in S\) when in state \(s\in S\) and taking action \(a\in A\); \(R\) is the reward function \(R:S \times A \mapsto \mathbb{R}\); \(0 < \gamma < 1\) is a discount factor; and \(H\) is the horizon time of the MDP, that is, the MDP runs for only \(H\) steps.\footnote{Any infinite horizon MDP with discounted rewards can be \(\epsilon\)-approximated by a finite horizon MDP using a horizon 
%%%%%\(H_\epsilon = [\log_{\gamma}(\epsilon(1-\gamma)/R_{\text{\it max}}]\), where \(R_\text{\it max} = \max_{s,a}|R(s,a)|\)~
%\(H_\epsilon = [\log_{\gamma}(\epsilon(1-\gamma))]/\max_{s,a} | R(s,a)|\)~
\(H_\epsilon = \frac{\log_{\gamma}(\epsilon(1-\gamma))}{\max_{s,a} | R(s,a)|}\)~
\cite{Jie2010Connection}.}  An agent selects its actions based on a policy \(\pi_\theta(\cdot|s)\), which is a distribution over all possible actions \(a\) in state \(s\) parameterized by \(\theta \in \mathbb{R}^n\). 

\paragraph{Policy Gradient Methods} One of the major approaches to deal with continuous control problems in MDPs is to apply a policy gradient method. In policy gradient methods, we are trying to use gradient ascent to optimize the following objective
\begin{equation}
\begin{split}   
J(\theta) & = E_{s \sim \rho^{\pi_{\theta}}}[V^{\pi_\theta}(s)]\\
& = \int_S  \rho^{\pi_{\theta}}(s)V^{\pi_\theta}(s)\diff s,
\end{split}
\label{eq:policy_gradient}
\end{equation}
where \(s\) is the state visited, and \(\rho^{\pi_\theta}(s)\) is the distribution over all states induced by executing policy \(\pi_\theta\). Many algorithms have been proposed to optimize this objective, including REINFORCE~\cite{Williams1992Simple}, GPOMDP~\cite{Baxter2001Infinite}, and Trust Region Policy Optimization (TRPO)~\cite{Schulman2015Trust}, where we collect a set of trajectory samples with the form \(\tau = \langle s_0, a_0, s_1, a_1, \ldots, s_H, a_H\rangle\) and use them to evaluate the gradient of \(J(\theta)\). It turns out that sometimes, it is beneficial to learn an additional value function \(Q(s,a)\) or \(V(s)\) to reduce the variance in estimating the gradient of \(J(\theta)\). This leads to a family of algorithms named ``actor-critic'' algorithms where the ``actor'' is the policy \(\pi\) and the ``critic'' is the value function. This family of algorithms includes the Stochastic Policy Gradient Theorem (SPG)~\cite{Sutton2000Policy}, the Deterministic Policy Gradient Theorem (DPG)~\cite{Silver2014Deterministic}, and so on. In addition, DDPG~\cite{Lillicrap2015Continuous} is an extention of DPG to the DRL setting by using a replay buffer to assist off-policy learning.

\subsection{Parameterized Action MDPs}

The MDP notation can be generalized to deal with parameterized tasks, e.g., actions with parameters. Here, instead of having just one set of actions, we have multiple sets of controls: a finite set of discrete actions \(A_d = \{a_1, a_2, \ldots, a_n\}\) and for each \(a \in A_d\), a set of continuous parameters \(X_a \subseteq R^{m_a}\). Thus, an action is a tuple \((a, x)\) in the joint action space, 
\[
A = \bigcup_{a\in A_d} \{(a, x) | x \in X_a\}.
\]
MDPs with this action space are called Parameterized Action MDPs (PAMDPs)~\cite{Masson2016Reinforcement}.

\subsection{Previous Work on Parameterized Action MDPs}
Tasks with parameterized actions have been a research topic in RL for a long time~\cite{Stone2005Keepaway}. \citeauthor{Zamani2012Symbolic} considered tasks with a set of discrete parameterized actions (\citeyear{Zamani2012Symbolic}). However, their algorithm which is based on Symbolic Dynamic Programming, is limited to MDPs with internal logical relations.

We adopt the Parameterized Action MDP setting from~\cite{Masson2016Reinforcement}. In their work, they train the policy in an alternative fashion. They first fix all the parameter policies, and hence induce an MDP with action set \(A\) of only discrete actions. Then they use Q-learning to learn a discrete policy in that MDP, and upon convergence, they fix the discrete policy, and start training the parameter policy. They show that this method can converge to local optima.

\citeauthor{Rachelson2009Timdppoly} used parameterized actions to deal with continuous time MDPs where the parameter for all the actions is the waiting time (\citeyear{Rachelson2009Timdppoly}). Thus, they have a unified parameter space. \citeauthor{Sharma2017Learning} did a similar approach where they extended TRPO to control the repetition of the action, that is, how many steps an action should execute (\citeyear{Sharma2017Learning}). They argued that the repetition times can be considered as a parameter for their original control signal. However, the repetition times are drawn from a fix set of integers, which is not a continuous signal.

The method that has the closest connection to our work is~\cite{Hausknecht2015Deep}, which extended the  DDPG to a parameterized action space. In this algorithm, the policy outputs all the parameters and all the discrete actions, and then selects the \((a, x)\) tuple with the highest Q-value. 

\section{Hierarchical Approaches in PAMDPs}
In this paper we propose a new, more natural architecture to generate actions for parameterized action tasks. In our algorithm, we have one neural network for the discrete policy and one neural network for the parameter policy. Our parameter policy \(\pi(x|s, a)\) takes two inputs, the state \(s\) and the discrete action \(a\) sampled from discrete action policy \(\pi(a|s)\). 
Then the joint action is given by 
\((a,x) \sim \pi(a, x|s) = \pi(a|s) \pi(x|s, a)\). Since the action \(a\) is known before we generate the parameters, we do not need the post processing step of determining which action tuple \((a,x)\) has the highest Q-values. And since the parameter policy knows the discrete action \(a\), the output size of parameter policy remains constant.

Previously, this architecture was not plausible in policy gradient methods due to the fact that we have to sample the discrete action in the middle of the forward pass, and the gradient cannot flow back to the discrete action policy in the backward pass due to the sampling operation. In this section, we describe two algorithms, Parameterized Action TRPO (PATRPO) and Parameterized Action SVG(0) (PASVG(0)) that solve this problem. 

Before we delve into the algorithms, we first introduce some notation. We use \(\pi_{\Theta}(a,x|s)\) to denote our overall policy, where \(a\) is the discrete action, \(x\) is the the parameter for that action, and \(\Theta\) is all parameters for the model. Our policy can be broken into two separate policies using conditional probability \(\pi_{\Theta}(a,x|s)=\pi^{c}_{\theta_x}(x|a,s)\pi^{d}_{\theta_a}(a|s)\), where \(\theta_a\) and \(\theta_x\) are the parameters for discrete action policy \(\pi^{d}(a|s)\) and continuous parameter policy \(\pi^{c}(x|a,s)\) respectively, and \(\Theta = [\theta_a, \theta_x]\). 

\subsection{Parameterized Actions TRPO}
Among all the policy gradient algorithms, TRPO and  DDPG achieve the best performance as they are able to optimize large neural network policies~\cite{Duan2016Benchmarking}. Thus, we consider how to apply these two algorithms in PAMDPs.

We first consider how to optimize our policy using TRPO's technique. In the TRPO, we are solving the following optimization problem:
\[
\begin{split}
& \text{maximize}_{\theta} \:\:L_{\theta'}(\theta) = E_{s \sim \rho_{\theta'}, a \sim \pi_{\theta'}}\bigg[\frac{\pi_{\theta}(a|s)}{\pi_{\theta'}(a|s)}Q_{\theta'}(s,a)\bigg]\\
& \text{subject to } \:\:\overline{\text{KL}}_{\theta'}(\theta) = E_{s \sim \rho_{\theta'}}[D_{\text{KL}}(\pi_{\theta'}(\cdot|s)||\pi_{\theta}(\cdot|s))] < \delta,
\end{split}
\]
where \(\theta'\) and \(\theta\) are the parameter vectors before and after each policy update respectively, and \(L\) is the surrogate loss. \(Q_{\theta}(s, a)\) indicates the Q-function fitted using the samples from policy parameterized by \(\theta\). The idea behind TRPO is to optimize the policy in a stable way such that the new policy distribution after each update will not be too different from the old one. This is achieved through the KL-divergence constraint between the policy distributions before and after the parameter update.

A similar idea has been explored before in the natural policy gradient~\cite{Kakade2002Natural}, where the objective function is replaced with linear approximation \(\frac{\partial L_{\theta'}(\theta)}{\partial \theta}(\theta-\theta')\) and the KL-divergence is replaced with a quadratic approximation \((\theta'-\theta)^{T}A(\theta'-\theta)\). The positive semidefinite matrix \(A\) in the quadratic term is the Hessian matrix of constraint, e.g., \(A = \frac{\partial^2}{\partial^2 \theta} \overline{\text{KL}}_{\theta'}(\theta) \). However, when the policy model becomes large,  \(A\) becomes very expensive to compute and store. What is special about TRPO is that it uses a Hessian-free optimization method~\cite{Martens2010Deep,Pearlmutter1994Fast} and conjugate gradient descent method to avoid the explicit formation of the Hessian matrix.  Therefore, TRPO only has a slight increase in the computation cost for optimizing large neural networks. 

To apply the TRPO in PAMDPs, we first write down the optimization problem using our notation, which is
\[
\begin{split}
& \text{maximize}_{\Theta} \:\: E_{s \sim \rho_{\Theta'}, (a,x) \sim \pi_{\Theta'}}\bigg[\frac{\pi_{\Theta}(a,x|s)}{\pi_{\Theta'}(a, x|s)}Q_{\Theta'}(s,a,x)\bigg]\\
& \text{subject to } \:\:E_{s \sim \rho_{\Theta'}}[D_{\text{KL}}(\pi_{\Theta'}(\cdot|s)||\pi_{\Theta}(\cdot|s))] < \delta
\end{split}
\]
The objective can be further expanded to
\[
E_{s \sim \rho_{\Theta'}, (a,x) \sim \pi_{\Theta'}}\bigg[\frac{\pi^{c}_{\theta_x}(x|a,s)\pi^{d}_{\theta_a}(a|s)}{\pi^{c}_{\theta'_x}(x|a,s)\pi^{d}_{\theta'_a}(a|s)}Q_{\Theta'}(s,a,x)\bigg]\\
\]
Notice that, in the objective function, the samples are collecting using \(\Theta'\) instead of \(\Theta\). Thus, in training time, we can just take the gradient of objective function w.r.t \(\Theta\) to achieve end-to-end training like normal supervised learning, and do not need to use any trick.

However, some changes are needed to meet the constraint of TRPO as there is no closed form solution for computing KL-divergence between two joint distributions. Here, we rewrite the KL-divergence constraint into conditional divergence using the chain rule. 
\[
\begin{split}
& E_{s \sim \rho_{\Theta'}}[D_{\text{KL}}(\pi_{\Theta'}(\cdot|s)||\pi_{\Theta}(\cdot|s))]\\
= & E_{s \sim \rho_{\Theta'}}\bigg[D_{\text{KL}}(\pi^{d}_{\theta'_a}(\cdot|s)||\pi^{d}_{\theta_a}(\cdot|s))\\
&+ E_{a \sim \pi^{d}_{\theta'_a}(a|s)}\big[D_{\text{KL}}(\pi^{c}_{\theta'_x}(\cdot|s,a)||\pi^{c}_{\theta_x}(\cdot|s,a))\big]\bigg]\\
= & E_{s \sim \rho_{\Theta'}}\bigg[D_{\text{KL}}(\pi^{d}_{\theta'_a}(\cdot|s)||\pi^{d}_{\theta_a}(\cdot|s))\bigg]\\
& + E_{s \sim \rho_{\Theta'}}E_{a \sim \pi^{d}_{\theta'_a}(a|s)}\bigg[D_{\text{KL}}(\pi^{c}_{\theta'_x}(\cdot|s,a)||\pi^{c}_{\theta_x}(\cdot|s,a))\bigg]\\
\end{split}
\] 
Thus, we can use samples to estimate both the objective function and KL-divergence. However, we notice that we can further reduce the variance of estimating the KL-divergence by using the analytical form of discrete action policy \(\pi(a|s)\). That is, the KL-divergence can be 
written as
\[
\begin{split}
&E_{s \sim \rho_{\Theta'}}\bigg[D_{\text{KL}}(\pi^{d}_{\theta'_a}(\cdot|s)||\pi^{d}_{\theta_a}(\cdot|s))\bigg]\\
&+ E_{s \sim \rho_{\Theta'}}\bigg[\pi(a|s) D_{\text{KL}}(\pi^{c}_{\theta'_x}(\cdot|s,a)||\pi^{c}_{\theta_x}(\cdot|s,a))\bigg]\\
\end{split}
\]
Using this form of constraint allows us to estimate the divergence between two policies even when we do not have samples for some discrete actions.

\subsection{Parameterized Actions SVG(0)}
Now we propose our second method based on the reparameterization trick.

One thing that makes the policy gradient methods special is that the samples we need to estimate the gradient come from the policy we are optimizing. That is, the objective usually takes the following form,
\[
E_{p_\theta(x)}[f(x)].
\]
We can write the gradient of expectation w.r.t \(\theta\) in this way:
\[
\begin{split}
\frac{\partial E_{p_\theta(x)}[f(x)]}{\partial \theta} & = \frac{\partial}{\partial \theta}\int_x  p_\theta(x)f(x)\diff x\\
& = \int_x  \frac{\partial p_\theta(x)}{\partial \theta} f(x) \diff x.\\
\end{split}
\]
Since we lost the term \(p(x)\) in the integral after we take the gradient, it's no longer an expectation, hence, we can no longer use samples from \(p(x)\) to estimate it. 

To solve this problem, people made the following changes to the gradient,
\[
\begin{split}
\frac{\partial E_{p_\theta(x)}[f(x)]}{\partial \theta} & = \int_x  \frac{\partial p_\theta(x)}{\partial \theta} f(x) \diff x\\
& = \int_x p(x) \bigg(\frac{1}{p(x)}\frac{\partial p_\theta(x)}{\partial \theta}\bigg) f(x) \diff x\\
& = E_{p_\theta(x)}\bigg[\frac{\partial \ln p_\theta(x)}{\partial \theta} f(x)\bigg]
\end{split}
\]
This trick is the foundation for most of the policy gradient methods in RL.
%This trick have been individually discovered in several fields and called \textit{score function estimator}~\cite{}, REINFORCE~\cite{}, and \textit{likelihood-ratio estimator}~\cite{} in different domains. 

Recently, another trick has been used to attack the same problem in the unsupervised learning community~\cite{Kingma2013Auto,Rezende2014Stochastic}. The idea is that a continuous random variable \(z\) can be obtained by first taking a noise variable \(\epsilon\) and then deterministically transforming it. For example, a gaussian random variable \(z \sim \mathcal{N}(\mu, \sigma^2)\) can be reparameterized into a noise random variable \(\epsilon \sim \mathcal{N}(0, 1)\) with a deterministc transformation \(g_{\mu, \sigma}(z) = \mu + \sigma \epsilon\). By applying this technique, we can optimize an expectation using samples from a noise distribution as follows
\[
E_{p_\theta(x)}[f(x)] = \int_x p_\theta(x) f(x) \diff x = \int_\epsilon p(\epsilon) f(g_\theta(\epsilon)) \diff \epsilon
\]
Then the gradient can be easily written as
\[
\frac{\partial E_{p_\theta(x)}[f(x)]}{\partial \theta}  = \int_\epsilon p(\epsilon) \bigg(\frac{\partial f}{\partial g} \frac{\partial g}{\partial \theta}\bigg) \diff \epsilon = E_{p(\epsilon)}\bigg[\frac{\partial f}{\partial g} \frac{\partial g}{\partial \theta}\bigg]
\]
This method has been successfully used  in Variational Autoencoders (VAE) for various works~\cite{Walker2016Uncertain,Sohn2015Learning}. It has also been applied to RL to train Stochastic Value Gradient (SVG) Learners~\cite{Heess2015Learning}. Recently, \citeauthor{Jang2016Categorical} (\citeyear{Jang2016Categorical}), \citeauthor{Maddison2016Concrete} (\citeyear{Maddison2016Concrete}) generalized the reparameterization trick to deal with discrete random variables with the \textit{Gumbel-Softmax} trick. In the discrete case, a random variable \(x\) can be drawn from a discrete distribution \(\{p(x_1), p(x_2), \ldots, p(x_n)\}\) by the Gumbel-Max trick~\cite{Maddison2014Sampling},
\[
x = \text{argmax}_i [ g_i + \ln p(x_i)]
\]
where \(g_i \sim \text{Gumbel}(0, 1)\). The Gumbel-Softmax trick replace the \(\text{argmax}\) operator in the above with a continuous differentiable \(\text{softmax}\) operator. With this change, we can now draw samples as
\[
x = \frac{\exp\big[\big((g_i + \ln p(x_i)\big)/t \big]}{\sum_{i=1}^n \exp\big[\big((g_i + \ln p(x_i)\big)/t \big]}
\]
where \(t\) is the ``temperature'' used to control the tradeoff between bias and variance. This trick has been applied to the RL setting as well, including imitation learning~\cite{Baram2017End} and multiagent RL~\cite{Mordatch2017Emergence}.

For our problem, the key observation is that the two steps of decision making in a parameterized action policy (choosing from a discrete action and then determining the parameters for it), is very much like the paradigm in VAE (\citeyear{Kingma2013Auto}). In the VAE setting, the encoder of the VAE takes a sample \(x\) from the dataset, and generates a latent variable \(z\) from it. Then the decoder takes \(z\) and reconstructs \(x\) out of it. For our situation, the discrete action policy first takes the state \(s\) as input and generates a discrete action \(a\), then determines the parameters \(x\) based on action \(a\) using the continuous parameter policy. Thus, we can roughly think of our discrete action policy and continuous parameter policy as the encoder and decoder in VAE respectively. 
%However, there are two key differences between the two models: 
%\begin{itemize}
%\item The objective function of VAE contains two relative separate objectives for encoder and decoder. The objective function for encoder is the KL-divergence between the approximate posterior and the true intractable posterior \(D_{\text{KL}}(q(z|x)||p(z|x))\), and the objective function for decoder is the log-likelihood of the reconstructed value \(x\) given \(z\). On contrary, in our setting, we have a unified objective function that applies to both discrete action policy and continuous parameter policy.
%\item The decoder of the VAE only takes \(z\) as the input, which is the output of the encoder. For our case, the continuous parameter policy takes both the action \(a\) from the discrete action policy and state \(s\) as input.
%\end{itemize}

\begin{figure}[t]
\begin{center}
\includegraphics[scale=0.35]{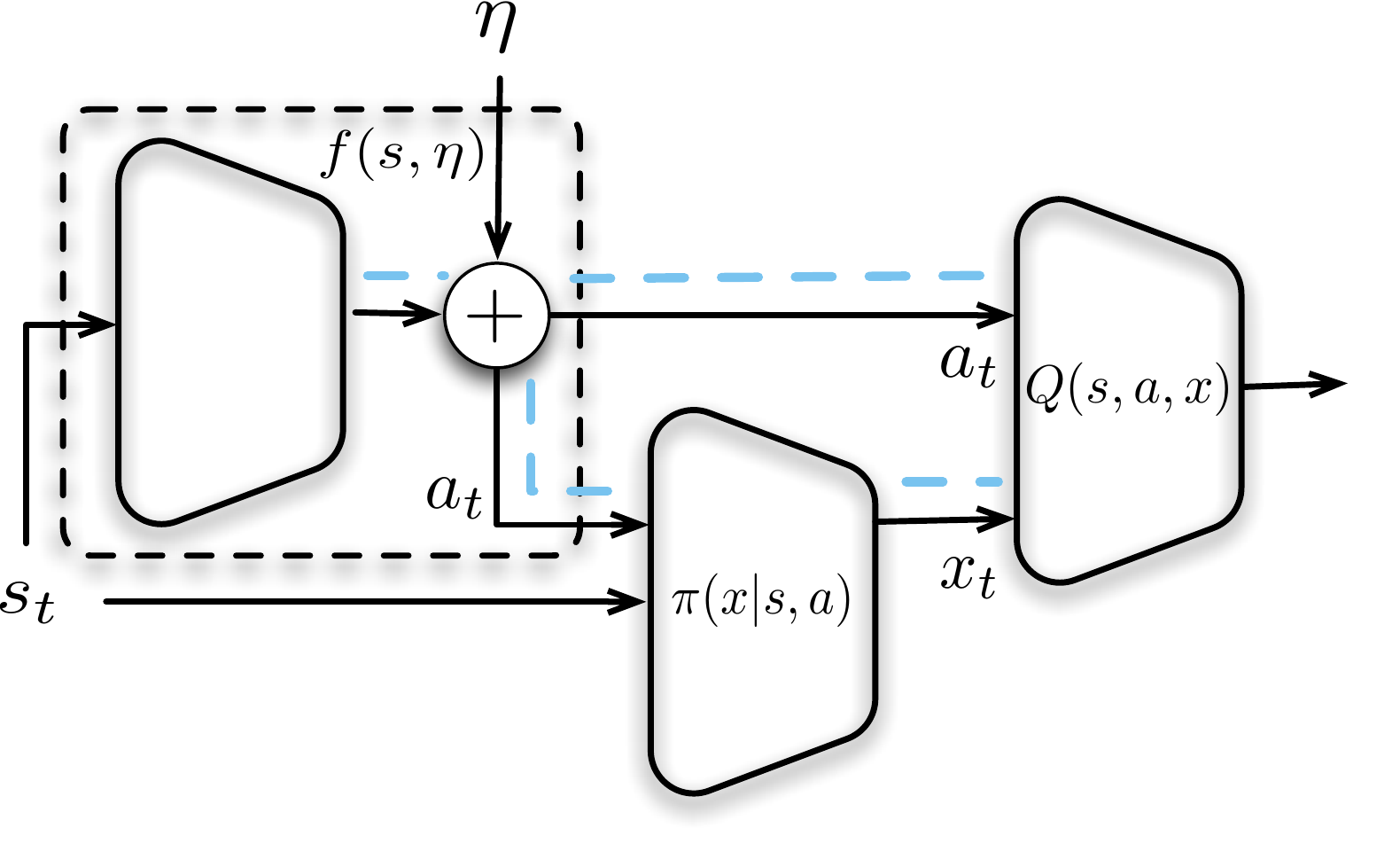}
\caption{The training flow of the PASVG(0) agent. The black lines indicate the forward pass of the training, and the dash lines indicate the backward pass of the training. The dash box marks the reparameterized policy \(f\).}
\label{fig:pasvg}
\end{center}
\end{figure}

We start with the objective function in (\ref{eq:policy_gradient}) and write it in parameterized action setting.
\[
\begin{split}
J(\Theta) & = \int_s \rho^{\pi_\Theta}(s) V^{\pi_\Theta}(s) ds\\
& = E_{s \sim \rho_{\Theta}}\bigg[\sum_a  \pi_{\Theta}(a, x|s) Q(s, a, x)\bigg]\\
& = E_{s \sim \rho_{\Theta}}\bigg[\sum_a \pi_{\theta_a}(a|s)  Q(s, a, \pi_{\theta_x} (x|s, a))\bigg]
\end{split}
\]
For the last step in the previous derivation, we use the DDPG formulation. Then we apply the reparameterization trick. Following the convention in~\cite{Heess2015Learning}, we use \(\eta\) to represent the auxiliary noise variable instead of \(\epsilon\) in the VAE setting.

\[
J(\Theta) = E_{s \sim \rho_{\Theta}}\bigg[\sum_{\eta} p(\eta)  Q(s, f_{\theta_a}(s, \eta), \pi_{\theta_x}(s, f_{\theta_a}(s, \eta)))\bigg]
\]
where \(a = f_{\theta_a}(s, \eta)\) is the discrete action policy after reparameterization. Then the gradient w.r.t \(\Theta\) is simply
\[
\frac{\partial J(\Theta)}{\partial \Theta} = E_{\rho_{\Theta}}E_{p(\eta)}\bigg[\frac{\partial}{\partial \Theta} Q(s, f_{\theta_a}(s, \eta), \pi_{\theta_x}(s, f_{\theta_a}(s, \eta)))\bigg]
\]
Since we are reparameterizing our stochastic policy for a 0-step value function (Q-function), similar to~\citeauthor{Heess2015Learning}'s method, thus we name our algorithm Parameterized Action SVG(0) (See Figure~\ref{fig:pasvg} for the training flow).

However, there is one critical difference between our work and~\citeauthor{Heess2015Learning}. In our work, we do not need to infer the noise variable since we are not using any dynamic model. To see this, we rewrite the gradient estimation using the Bayes' rule, following the method from~\cite{Heess2015Learning}, 
\begin{multline}
\frac{\partial J(\Theta)}{\partial \Theta} = E_{\rho_{\Theta}}E_{\pi(a,x|s)}E_{p(\eta|a, x, s)}\\
\bigg[\frac{\partial}{\partial \Theta} Q(s, f_{\theta_a}(s, \eta), \pi_{\theta_x}(s, f_{\theta_a}(s, \eta)))\bigg]
\end{multline}

\citeauthor{Heess2015Learning} use this method to infer the noise \(\zeta\) of their reparameterized approximate dynamic model \(s' = g(s, a, \zeta)\). Thus, they need to learn the \(p(\zeta|s, a, s')\) which is similar to \(p(\eta|a, x, s)\) in our case. However, for us, we use the sample \(\eta\) and generate \(a, x\) from it. Hence, we do not need to learn the model \(p(\eta|a,x,s)\). Instead, we can just record the value of \(\eta\) when we are collecting the training samples.

The last part of the algorithm is to make the gradient estimation not depend on the samples collected by \(\pi(a,x|s)\), as the policy is constantly changing. We use the replay buffer from DDPG to solve this issue and turn our algorithm into an off-policy algorithm to improve sample efficiency.

%\begin{algorithm}
%\SetAlgoLined
%\KwResult{Write here the result }
%	initialization\;
%	Given empty experience database \(D\)\\
%	\For{t = 0 to \(\infty\)}{
%		Get discrete action \(a = f(s, \eta_a), \eta_a \sim \rho(\eta_a)\)\\
%		Get the corresponding parameters \(x = g(s, a, \eta_x), \eta_x \sim \rho(\eta_x)\)\\
%		Apply the control \((a, x)\)\\
%		Observe \(r, s'\)\\
%		Insert \((s, \eta_a, a, \eta_x, x, r, s')\) into \(D\)\\
%		Take a minibatch of sample from \(D\) and train \(Q(s, a, x)\) on them (Need to expand here)\\		
%	}
% \caption{Parameterized Action SVG(0)}
% \end{algorithm}

\section{Experiments}

\begin{figure}[t]
\frame{\includegraphics[scale=0.229]{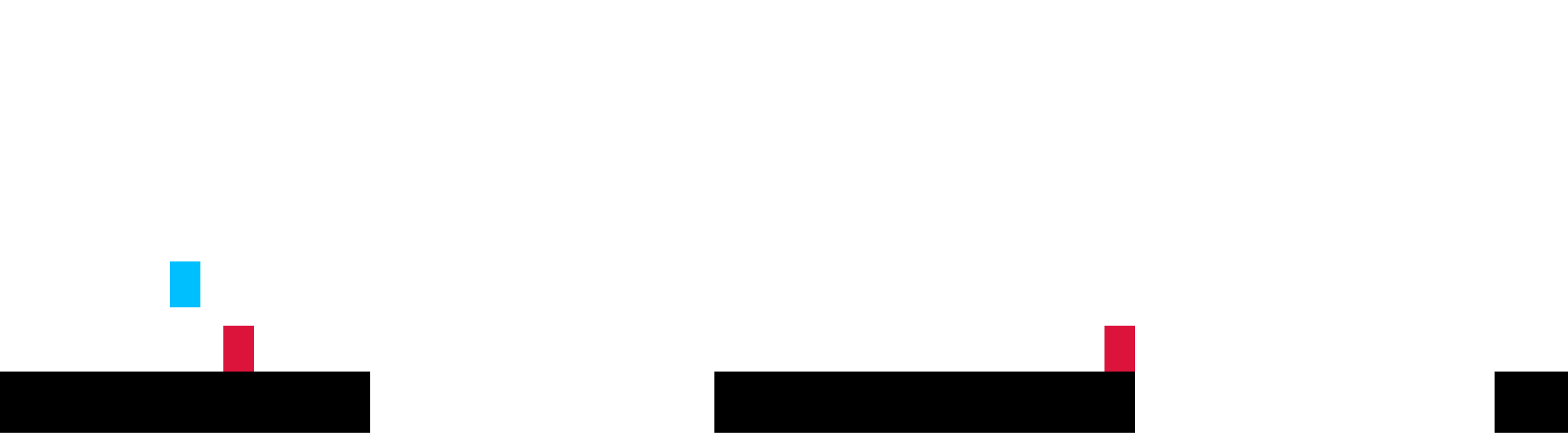}}
\caption{Platform domain}
\label{fig:platform}
\end{figure}

We conducted our experiments using the Platform domain from~\cite{Masson2016Reinforcement} (See Figure~\ref{fig:platform}). In this domain, we control the agents (cyan block) to jump across several platforms while avoiding enemies (red blocks) and falling off the platforms. This domain has three discrete actions to choose from: run, jump, and leap. A jump moves the agent over its enemies, while a leap propels the agent to the next platform. Each of the actions take one parameter which determines the speed along the x-axis. More details of the domain can be found in the original paper. 

We implemented the Parameterization Action DDPG\footnote[2]{This is the DDPG algorithm for parametereized action spaces, not to be confused with the DDPG algorithm from~\cite{Lillicrap2015Continuous} for continuous control.} (PADDPG) algorithm from~\cite{Hausknecht2015Deep} as our comparison baseline which is considered as the state of the art. Specifically, we implemented the PADDPG algorithm following the settings and parameters from the original paper except for the size of the hidden layers. In the original paper, PADDPG used a huge network with four hidden layers with size \{1024, 512, 256, 128\}, which is rare in DRL community for tasks with continuous signals. We followed the DDPG paper~\cite{Lillicrap2015Continuous}, which used two hidden layers with sizes \{400, 300\} for the neural networks. We also implemented their invert-gradient trick, as they claimed that this was the only way to make the learning work in a bounded parameter space. 

For our PATRPO agent, we adopted the setting from~\cite{Duan2016Benchmarking}, where we had three hidden layers with sizes \{200, 100, 50\} for the policies. We used ReLU for activation, and Softmax and Tanh for the output layers of the discrete action policy and continuous parameter policy respectively. For our PASVG(0) agent, we also used neural networks with two hidden layers of sizes \{400, 300\} and ReLU for activation. For the output layer, we used Gumbel Softmax for the discrete action policy and Tanh for the continuous parameter policy.

We trained all the agents using 100 epochs with 10000 samples per epoch. For the online method, we had a replay buffer with size \(10^7\) and we did not start the training until we had \(10^4\) samples in the replay buffer, which is a standard setting in DRL experiments. We used 0.005 as the step size for PATRPO agent and \(10^{-3}\) and \(10^{-5}\) as the learning rate for the value function and policies respectively for the PASVG(0) agents. We fixed the temperature to 1.0 for the Gumbel-Softmax layer and kept it for the entire training process. 

\begin{figure}[t]
\includegraphics[width=3.2in]{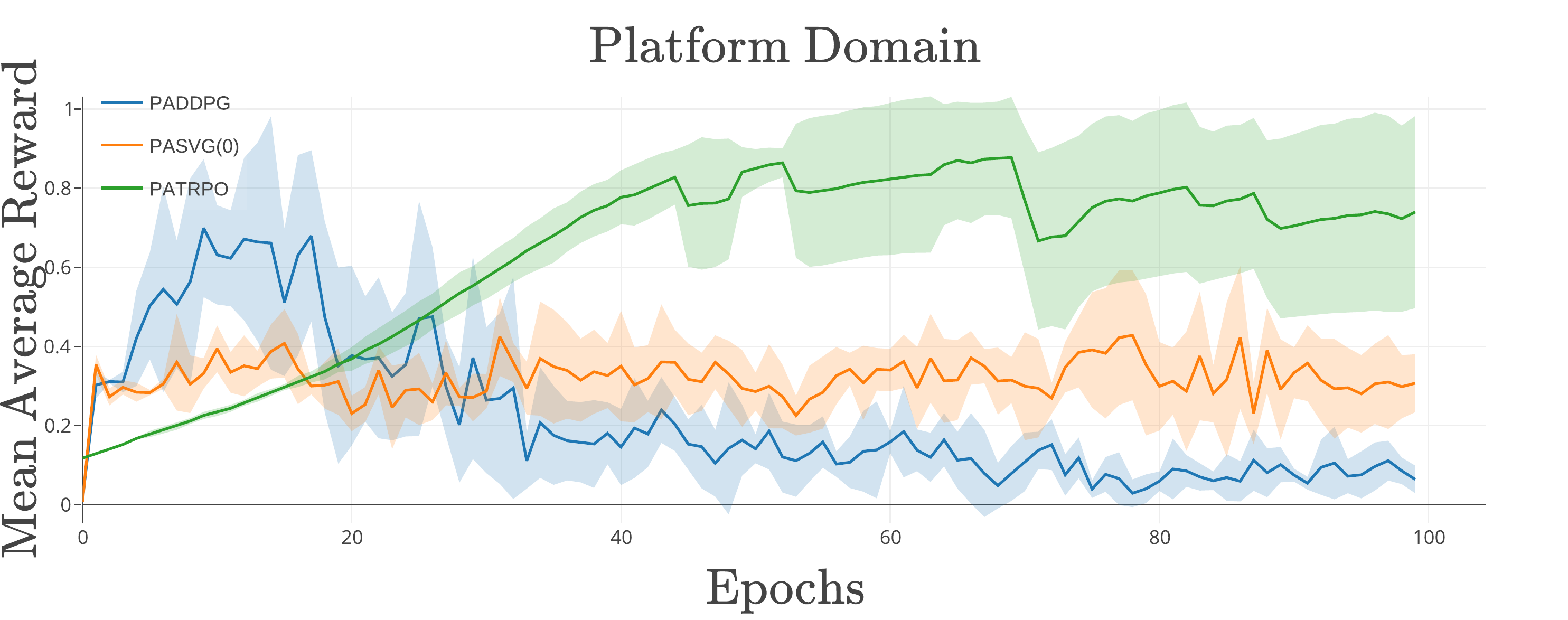}
\caption{Comparison on Platform domain of three learners. The x-axis shows the training epochs. The y-axis shows the average reward. Solid lines are average value over five random seeds. Shaded regions are standard deviation.}
\label{fig:platform_comparison}
\end{figure}

The experiment results are shown in Figure~\ref{fig:platform_comparison}. The plot of PADDPG is very interesting: we found that it can learn to successfully finish the game at an early stage of learning, but would quickly unlearn that policy and converge to something else. 

%We hypothesis that this is due to the fact in the training process, its Q-network take all the action and parameter values as input and do not distinguish them, thus fail to track the true best policy.

We then noted that, although we are using Tanh to bound the output of our parameter policy, which corresponded to the squash-gradient method in~\cite{Hausknecht2015Deep}, we managed to make it work for our methods, which suggests that there are more training options than the invert-gradient method suggested by~\cite{Hausknecht2015Deep}. Our PATRPO method achieved the best performance among all three learners, and unlike PADDPG learner, it maintained its performance after obtaining its best learned policy. The PASVG(0) learner converged to a local optimum with average reward of around 0.4. By examining the game, we found that this corresponded to avoiding the first enemy but failing to land on the second platform. One of the possible reasons was that the learner was conducting joint-learning, which is very similar to cooperative multiagent learning, and thus may converge to a local optima.

%However, the PATRPO learner avoid this by using a small step size, which limit the co-adaption in this multiagent learning like scenarios.

\begin{figure}[t]
\includegraphics[width=3.2in]{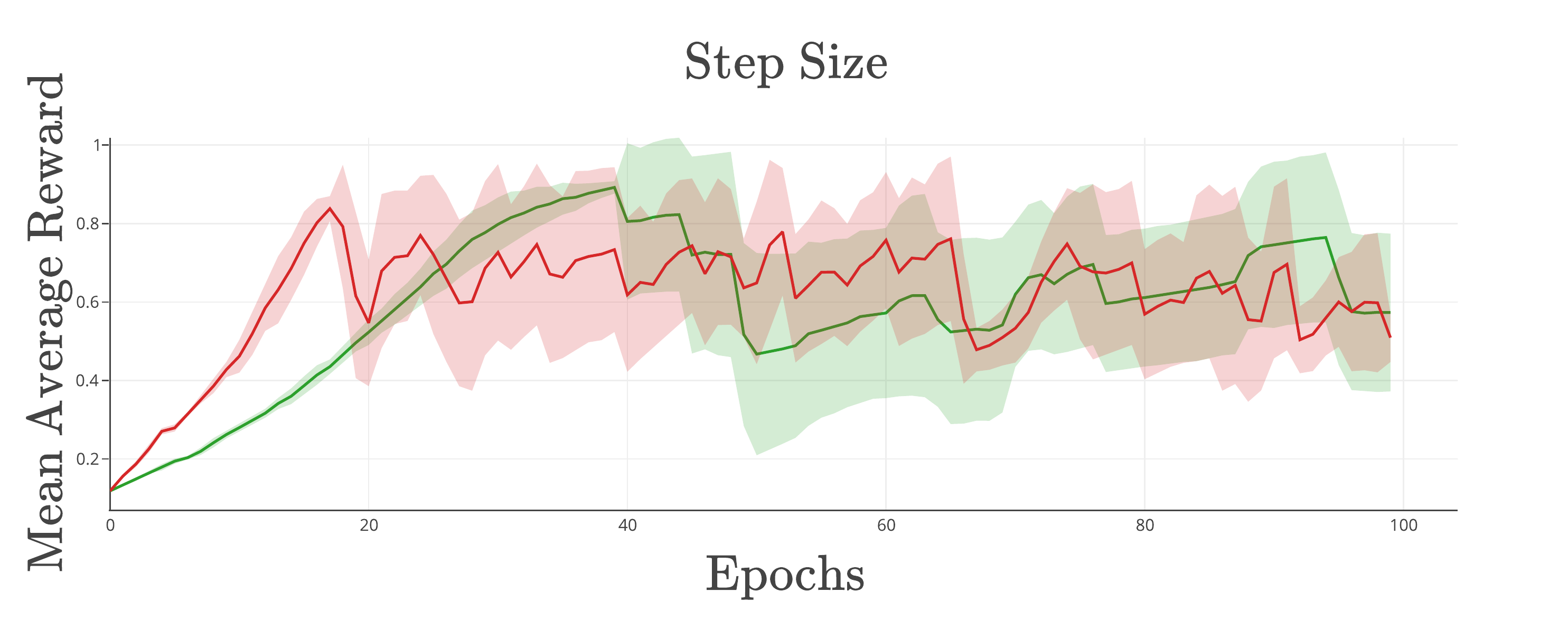}
\caption{Different step size parameters for PATRPO agents. \(\delta = 0.05\) in red and \(\delta = 0.01\) in green.}
\label{fig:step_size}
\end{figure}

We further investigated this joint-learning issue by trying two different step sizes for the PATRPO agent. Figure~\ref{fig:step_size} shows the result of using larger step size parameters. As we can see, both of the agents can achieve a good performance in relatively short period of time with much lower variance. But once they learn the optimal policy, their performance starts to drop and the variance becomes much larger. However, PATRPO still manages to maintain a decent performance which is far better than the PADDPG algorithm. This shows that a more stable method is desirable for learning in the parameterized action space.

\begin{figure}[t]
\includegraphics[width=3.2in]{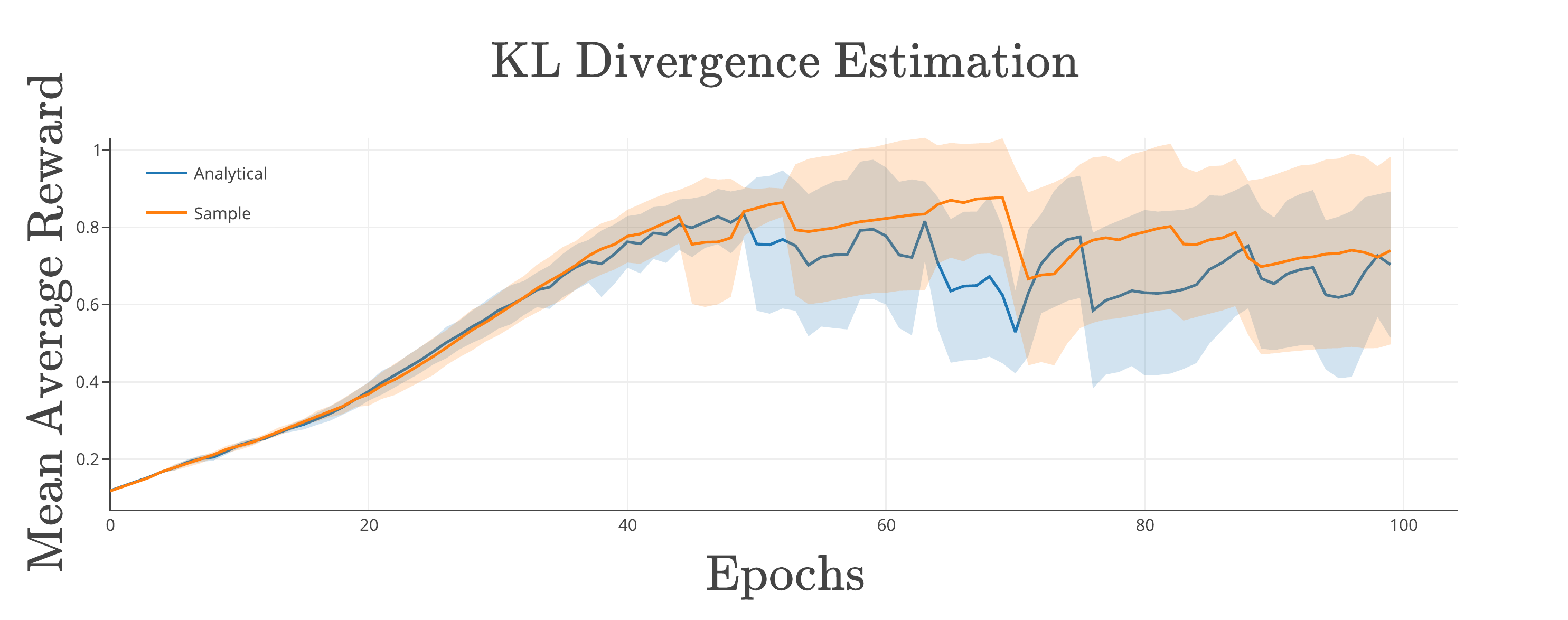}
\caption{Comparison on the Platform domain for different KL-Divergence estimation methods.}
\label{fig:kl_div}
\end{figure}

Last, we conducted an experiment using different techniques for estimating the KL divergence in PATRPO. The experiment as illustrated in Figure~\ref{fig:kl_div} showed that none of them makes much of a difference in this small domain. 

\begin{figure}[t]
\begin{center}
\includegraphics[scale=0.4, angle=90]{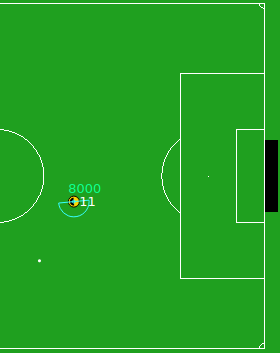}
\end{center}
\caption{An example of Half Field Offense Domain, with no goalie.}
\label{fig:hfo}
\end{figure}

We also tested our algorithm in the HFO domain introduced by~\cite{Hausknecht2015Deep}. In this domain (Figure~\ref{fig:hfo}), we controlled an agent to score a goal. For the simplicity, we did not have a goalie. We had three actions in this domain, \textbf{dash} with parameters {\it power} and {\it direction}, \textbf{turn} with parameter {\it direction} and \textbf{kick} with parameters {\it power} and {\it direction}. Thus, different actions required different numbers of parameters. For our agents, if we outputed more parameters than we actually needed, we just took the first part of the output and ignored the remainder. This domain had a 59-dimensional state space, which was much larger compared to the platform domain, and thus we trained our agents using larger neural networks and with more samples. Due to time constraints, we only trained our PATRPO agent and PADDPG in this domain for 100 epochs with 50000 steps per epoch. We used three hidden layers with size \{400, 300, 200\} for both PATRPO and PADDPG agents. 

\begin{figure}[t]
\includegraphics[width=3.2in]{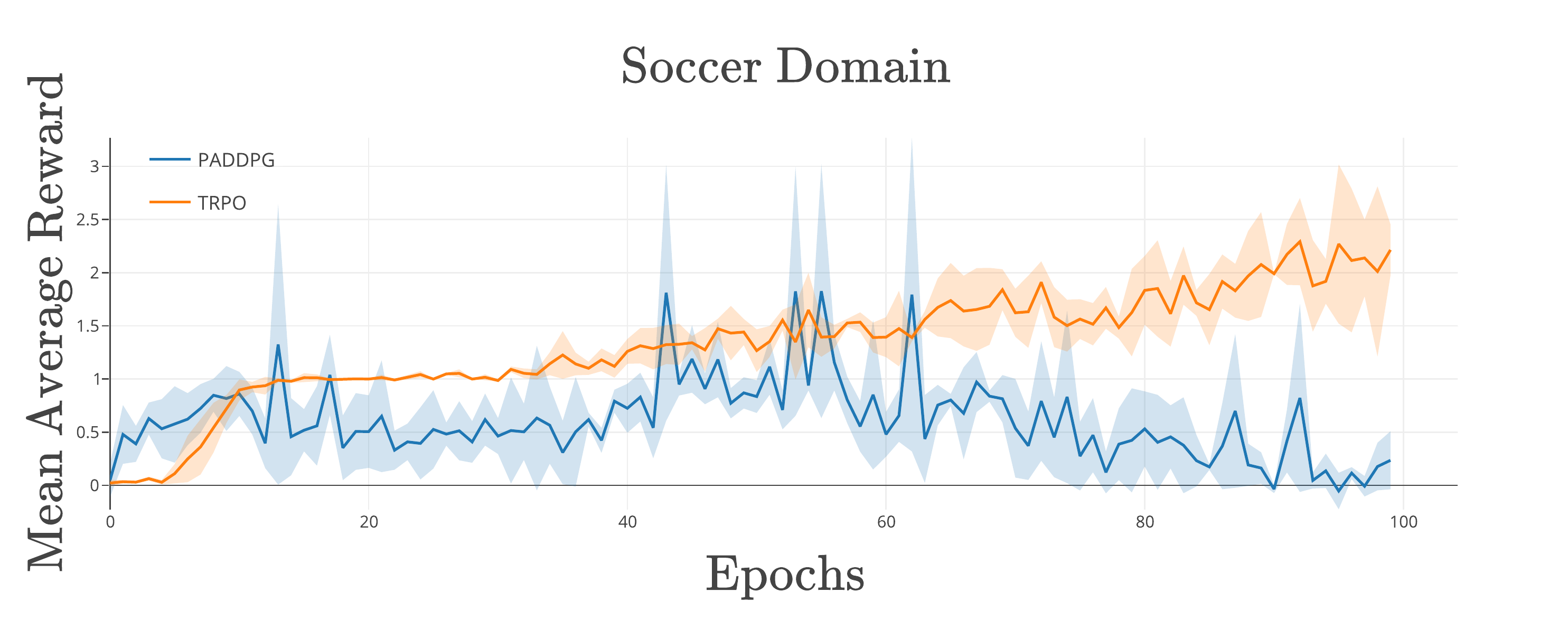}
\caption{Comparison on Soccer domain for PATRPO and PADDPG agents on three different random seeds.}
\label{fig:soccer}
\end{figure}

Figures~\ref{fig:soccer} shows the result on this domain. As we can see, again, the PATRPO agent achieved stable performance in this domain while PADDPG demonstrated a large variance in its performance. We also note that the performance of the PADDPG algorithm is far worse than in the original paper. One of the possible reasons for this is due to the difference in the neural network sizes. But since our PATRPO agent can achieve stable learning in this domain with a much smaller neural network, this suggests that a large neural network is not necessary in this domain.

\section{Conclusion and Future Work}
%
% We give detailed background information on relevant topics.
% We present two algorithms for learning effective control in parameterized action space.
% We use the analytical form of discrete action policy so as to further reduce the variance of estimating the KL-divergence.
% We incorporate the reparameterization trick in the second method

We presented two algorithms for learning effective control in parameterized action space. We demonstrated that our method can learn better policy in these setting compared to PADDPG method. However, we found that learning could be unstable due to the joint-learning between the discrete action policy and parameter policy. An interesting future direction would be to find more stable methods for this domain. 
We would like to study these methods in the context of more complex domains (in soccer for example) particularly involving more agents.

\section{Acknowledgments}
The research in this paper was conducted with the support of research infrastructure developed under NSF grant 1727303.
\bibliography{parameterized}

\begin{thebibliography}{}

\bibitem[\protect\citeauthoryear{Baram \bgroup et al\mbox.\egroup
  }{2017}]{Baram2017End}
Baram, N.; Anschel, O.; Caspi, I.; and Mannor, S.
\newblock 2017.
\newblock End-to-end differentiable adversarial imitation learning.
\newblock In {\em Proceedings of the 34th International Conference on Machine
  Learning}, volume~70,  390--399.

\bibitem[\protect\citeauthoryear{Baxter and
  Bartlett}{2001}]{Baxter2001Infinite}
Baxter, J., and Bartlett, P.~L.
\newblock 2001.
\newblock Infinite-horizon policy-gradient estimation.
\newblock {\em Journal of Artificial Intelligence Research}  319--350.

\bibitem[\protect\citeauthoryear{Duan \bgroup et al\mbox.\egroup
  }{2016}]{Duan2016Benchmarking}
Duan, Y.; Chen, X.; Houthooft, R.; Schulman, J.; and Abbeel, P.
\newblock 2016.
\newblock Benchmarking deep reinforcement learning for continuous control.
\newblock In {\em Proceedings of The 33rd International Conference on Machine
  Learning},  1329--1338.

\bibitem[\protect\citeauthoryear{Hausknecht and
  Stone}{2015}]{Hausknecht2015Deep}
Hausknecht, M., and Stone, P.
\newblock 2015.
\newblock Deep reinforcement learning in parameterized action space.
\newblock {\em arXiv preprint arXiv:1511.04143}.

\bibitem[\protect\citeauthoryear{Heess \bgroup et al\mbox.\egroup
  }{2015}]{Heess2015Learning}
Heess, N.; Wayne, G.; Silver, D.; Lillicrap, T.; Erez, T.; and Tassa, Y.
\newblock 2015.
\newblock Learning continuous control policies by stochastic value gradients.
\newblock In {\em Advances in Neural Information Processing Systems},
  2944--2952.

\bibitem[\protect\citeauthoryear{Jang, Gu, and
  Poole}{2016}]{Jang2016Categorical}
Jang, E.; Gu, S.; and Poole, B.
\newblock 2016.
\newblock Categorical reparameterization with gumbel-softmax.
\newblock {\em arXiv preprint arXiv:1611.01144}.

\bibitem[\protect\citeauthoryear{Jie and Abbeel}{2010}]{Jie2010Connection}
Jie, T., and Abbeel, P.
\newblock 2010.
\newblock On a connection between importance sampling and the likelihood ratio
  policy gradient.
\newblock In {\em Advances in Neural Information Processing Systems},
  1000--1008.

\bibitem[\protect\citeauthoryear{Kakade}{2002}]{Kakade2002Natural}
Kakade, S.~M.
\newblock 2002.
\newblock A natural policy gradient.
\newblock In {\em Advances in neural information processing systems},
  1531--1538.

\bibitem[\protect\citeauthoryear{Kingma and Welling}{2013}]{Kingma2013Auto}
Kingma, D.~P., and Welling, M.
\newblock 2013.
\newblock Auto-encoding variational bayes.
\newblock {\em arXiv preprint arXiv:1312.6114}.

\bibitem[\protect\citeauthoryear{Lillicrap \bgroup et al\mbox.\egroup
  }{2015}]{Lillicrap2015Continuous}
Lillicrap, T.~P.; Hunt, J.~J.; Pritzel, A.; Heess, N.; Erez, T.; Tassa, Y.;
  Silver, D.; and Wierstra, D.
\newblock 2015.
\newblock Continuous control with deep reinforcement learning.
\newblock {\em arXiv preprint arXiv:1509.02971}.

\bibitem[\protect\citeauthoryear{Maddison, Mnih, and
  Teh}{2016}]{Maddison2016Concrete}
Maddison, C.~J.; Mnih, A.; and Teh, Y.~W.
\newblock 2016.
\newblock The concrete distribution: A continuous relaxation of discrete random
  variables.
\newblock {\em arXiv preprint arXiv:1611.00712}.

\bibitem[\protect\citeauthoryear{Maddison, Tarlow, and
  Minka}{2014}]{Maddison2014Sampling}
Maddison, C.~J.; Tarlow, D.; and Minka, T.
\newblock 2014.
\newblock A* sampling.
\newblock In {\em Advances in Neural Information Processing Systems 27},
  3086--3094.

\bibitem[\protect\citeauthoryear{Martens}{2010}]{Martens2010Deep}
Martens, J.
\newblock 2010.
\newblock Deep learning via hessian-free optimization.
\newblock In {\em Proceedings of the 27th International Conference on Machine
  Learning (ICML-10)},  735--742.

\bibitem[\protect\citeauthoryear{Masson, Ranchod, and
  Konidaris}{2016}]{Masson2016Reinforcement}
Masson, W.; Ranchod, P.; and Konidaris, G.
\newblock 2016.
\newblock Reinforcement learning with parameterized actions.
\newblock In {\em AAAI},  1934--1940.

\bibitem[\protect\citeauthoryear{Mnih \bgroup et al\mbox.\egroup
  }{2015}]{Mnih2015Human}
Mnih, V.; Kavukcuoglu, K.; Silver, D.; Rusu, A.~A.; Veness, J.; Bellemare,
  M.~G.; Graves, A.; Riedmiller, M.; Fidjeland, A.~K.; Ostrovski, G.; et~al.
\newblock 2015.
\newblock Human-level control through deep reinforcement learning.
\newblock {\em Nature} 518(7540):529--533.

\bibitem[\protect\citeauthoryear{Mordatch and
  Abbeel}{2017}]{Mordatch2017Emergence}
Mordatch, I., and Abbeel, P.
\newblock 2017.
\newblock Emergence of grounded compositional language in multi-agent
  populations.
\newblock {\em CoRR} abs/1703.04908.

\bibitem[\protect\citeauthoryear{Pearlmutter}{1994}]{Pearlmutter1994Fast}
Pearlmutter, B.~A.
\newblock 1994.
\newblock Fast exact multiplication by the hessian.
\newblock {\em Neural computation} 6(1):147--160.

\bibitem[\protect\citeauthoryear{Rachelson, Fabiani, and
  Garcia}{2009}]{Rachelson2009Timdppoly}
Rachelson, E.; Fabiani, P.; and Garcia, F.
\newblock 2009.
\newblock Timdppoly: An improved method for solving time-dependent mdps.
\newblock In {\em Tools with Artificial Intelligence, 2009. ICTAI'09. 21st
  International Conference on},  796--799.
\newblock IEEE.

\bibitem[\protect\citeauthoryear{Rezende, Mohamed, and
  Wierstra}{2014}]{Rezende2014Stochastic}
Rezende, D.~J.; Mohamed, S.; and Wierstra, D.
\newblock 2014.
\newblock Stochastic backpropagation and approximate inference in deep
  generative models.
\newblock In {\em Proceedings of the 31st International Conference on Machine
  Learning (ICML-14)},  1278--1286.

\bibitem[\protect\citeauthoryear{Schulman \bgroup et al\mbox.\egroup
  }{2015}]{Schulman2015Trust}
Schulman, J.; Levine, S.; Abbeel, P.; Jordan, M.; and Moritz, P.
\newblock 2015.
\newblock Trust region policy optimization.
\newblock In {\em Proceedings of the 32nd International Conference on Machine
  Learning (ICML-15)},  1889--1897.

\bibitem[\protect\citeauthoryear{Sharma, Lakshminarayanan, and
  Ravindran}{2017}]{Sharma2017Learning}
Sharma, S.; Lakshminarayanan, A.~S.; and Ravindran, B.
\newblock 2017.
\newblock Learning to repeat: Fine grained action repetition for deep
  reinforcement learning.
\newblock {\em arXiv preprint arXiv:1702.06054}.

\bibitem[\protect\citeauthoryear{Silver \bgroup et al\mbox.\egroup
  }{2014}]{Silver2014Deterministic}
Silver, D.; Lever, G.; Heess, N.; Degris, T.; Wierstra, D.; and Riedmiller, M.
\newblock 2014.
\newblock Deterministic policy gradient algorithms.
\newblock In {\em ICML}.

\bibitem[\protect\citeauthoryear{Silver \bgroup et al\mbox.\egroup
  }{2016}]{Silver2016Mastering}
Silver, D.; Huang, A.; Maddison, C.~J.; Guez, A.; Sifre, L.; Van Den~Driessche,
  G.; Schrittwieser, J.; Antonoglou, I.; Panneershelvam, V.; Lanctot, M.;
  et~al.
\newblock 2016.
\newblock Mastering the game of go with deep neural networks and tree search.
\newblock {\em Nature} 529(7587):484--489.

\bibitem[\protect\citeauthoryear{Sohn, Lee, and Yan}{2015}]{Sohn2015Learning}
Sohn, K.; Lee, H.; and Yan, X.
\newblock 2015.
\newblock Learning structured output representation using deep conditional
  generative models.
\newblock In {\em Advances in Neural Information Processing Systems},
  3483--3491.

\bibitem[\protect\citeauthoryear{Stone \bgroup et al\mbox.\egroup
  }{2006}]{Stone2005Keepaway}
Stone, P.; Kuhlmann, G.; Taylor, M.~E.; and Liu, Y.
\newblock 2006.
\newblock Keepaway soccer: From machine learning testbed to benchmark.
\newblock In Noda, I.; Jacoff, A.; Bredenfeld, A.; and Takahashi, Y., eds.,
  {\em {R}obo{C}up-2005: Robot Soccer World Cup {IX}}, volume 4020. Berlin:
  Springer Verlag.
\newblock  93--105.

\bibitem[\protect\citeauthoryear{Sutton \bgroup et al\mbox.\egroup
  }{2000}]{Sutton2000Policy}
Sutton, R.~S.; McAllester, D.~A.; Singh, S.~P.; and Mansour, Y.
\newblock 2000.
\newblock Policy gradient methods for reinforcement learning with function
  approximation.
\newblock In {\em Advances in neural information processing systems},
  1057--1063.

\bibitem[\protect\citeauthoryear{Vinyals \bgroup et al\mbox.\egroup
  }{2017}]{Vinyals2017Starcraft}
Vinyals, O.; Ewalds, T.; Bartunov, S.; Georgiev, P.; Vezhnevets, A.~S.; Yeo,
  M.; Makhzani, A.; K{\"u}ttler, H.; Agapiou, J.; Schrittwieser, J.; et~al.
\newblock 2017.
\newblock Starcraft ii: A new challenge for reinforcement learning.
\newblock {\em arXiv preprint arXiv:1708.04782}.

\bibitem[\protect\citeauthoryear{Walker \bgroup et al\mbox.\egroup
  }{2016}]{Walker2016Uncertain}
Walker, J.; Doersch, C.; Gupta, A.; and Hebert, M.
\newblock 2016.
\newblock An uncertain future: Forecasting from static images using variational
  autoencoders.
\newblock In {\em European Conference on Computer Vision},  835--851.
\newblock Springer.

\bibitem[\protect\citeauthoryear{Williams}{1992}]{Williams1992Simple}
Williams, R.~J.
\newblock 1992.
\newblock Simple statistical gradient-following algorithms for connectionist
  reinforcement learning.
\newblock {\em Machine learning} 8(3-4):229--256.

\bibitem[\protect\citeauthoryear{Zamani \bgroup et al\mbox.\egroup
  }{2012}]{Zamani2012Symbolic}
Zamani, Z.; Sanner, S.; Fang, C.; et~al.
\newblock 2012.
\newblock Symbolic dynamic programming for continuous state and action mdps.
\newblock In {\em AAAI}.

\end{thebibliography}
\bibliographystyle{aaai}

\end{document}